\pdfoutput=1

\documentclass[11pt]{article}

\usepackage{times}
\usepackage{latexsym}

\usepackage[T1]{fontenc}

\usepackage[utf8]{inputenc}

\usepackage{inconsolata}
\usepackage{authblk}
\usepackage{acl2023}
\usepackage{booktabs}
\usepackage{graphicx}

\usepackage{microtype}
\usepackage{dsfont}
\usepackage{amsmath}
\usepackage{amssymb}
\usepackage{hyperref}
\usepackage{caption}
\usepackage{subcaption}
\usepackage{rotating}

\title{RepMatch: Quantifying Cross-Instance Similarities \\in Representation Space}

\usepackage[T1]{fontenc}
\usepackage[utf8]{inputenc}

\author{
    \textbf{Mohammad Reza Modarres$^1$} ~
    \textbf{Sina Abbasi$^1$} ~
    \textbf{Mohammad Taher Pilehvar$^2$} \\
    $^1$Tehran Institute for Advanced Studies, Khatam University, Iran \\
    $^2$Cardiff University, United Kingdom\\
    \texttt{m.modares401@khatam.ac.ir} ~
    \texttt{s.abbasi401@khatam.ac.ir} ~
    \texttt{mp792@cam.ac.uk}
}

\begin{document}
\maketitle
\begin{abstract}
Advances in dataset analysis techniques have enabled more sophisticated approaches to analyzing and characterizing training data instances, often categorizing data based on attributes such as ``difficulty''. 
In this work, we introduce RepMatch, a novel method that characterizes data through the lens of similarity.
RepMatch quantifies the similarity between subsets of training instances by comparing the knowledge encoded in models trained on them, overcoming the limitations of existing analysis methods that focus solely on individual instances and are restricted to within-dataset analysis.
Our framework allows for a broader evaluation, enabling similarity comparisons across arbitrary subsets of instances, supporting both dataset-to-dataset and instance-to-dataset analyses. We validate the effectiveness of RepMatch across multiple NLP tasks, datasets, and models. Through extensive experimentation, we demonstrate that RepMatch can effectively compare datasets, identify more \textit{representative} subsets of a dataset (that lead to better performance than randomly selected subsets of equivalent size), and uncover heuristics underlying the construction of some challenge datasets.

\end{abstract}

\section{Introduction}

Contemporary machine learning models are deeply influenced by the datasets on which they are trained. 
The characteristics of a dataset, encompassing the quality and diversity of its instances, are critical in shaping a model’s learning effectiveness and its capability to generalize. 
Recent advancements in the field have led to the development of methodologies that facilitate the analysis and categorization of data instances based on specific attributes, notably ``difficulty'' \citep{ethayarajh2022understanding, siddiqui2022metadata, swayamdipta-etal-2020-dataset}, as well as other attributes, such as noisiness, atypicality, prototypicality, and distributional outliers \citep{siddiqui2022metadata}. 
These methodologies often involve ranking or categorizing training instances based on specific attributes to identify types that may require specialized processing or treatment. One possible application is to detect mislabeled or noisy examples; removing these from the training data can lead to more effective training \cite{mirzasoleiman2020coresets, Pleiss2020}. 
Additionally, these analysis methods are instrumental in studying and uncovering dataset artifacts \citep{gardner2021competency, ethayarajh2022understanding}.

Despite their contributions, existing methods are often limited by their focus on individual instances without the capacity to evaluate subsets of data as a whole. 
Moreover, they are generally confined to analyses within a single dataset, lacking the ability to compare across different datasets or perform comprehensive cross-dataset evaluations.

In response to these limitations, we introduce a novel technique, called RepMatch, that offers a refined lens for the analysis: quantifying the similarities between subsets of training instances from the perspective of the models trained on them.
Specifically, we measure the similarity between two subsets, 
$\mathcal{S} \subseteq \mathcal{D}$ and $\mathcal{S}^\prime \subseteq \mathcal{D}^\prime$ of the training datasets $\mathcal{D}$ and $\mathcal{D}^\prime$ (where $\mathcal{D}$ and $\mathcal{D}^\prime$ could be the same dataset), by comparing the models trained exclusively on each subset.
The subsets are deemed similar if the representation space learned by the model trained on 
$\mathcal{S}$ closely aligns with that learned by the model trained on $\mathcal{S}^\prime$.
This reformulation overcomes previous limitations by enabling the analysis and evaluation of similarities among arbitrary subsets of instances—from individual examples to entire datasets—across varied sources.
Specifically, RepMatch facilitates:

\begin{itemize}
    \item \textbf{Dataset-dataset analysis}:
    Compare similarities in task and dataset characteristics from a model's perspective, both within and beyond their original domains.
    
    \item \textbf{Instance-dataset analysis}:
    Identify the most \textit{representative} instances from the target dataset (or others), using which a more effective training can be performed. Similarly, identify those with the \textit{least} information, suggesting out-of-distribution or noisy outliers.

\end{itemize}

Comparing two models—particularly modern, heavily parameterized ones with expansive weight matrices—presents a significant challenge. 
To address this, we constrain the set of trainable parameters, i.e., the updates in the representation space, by leveraging Low-rank Adaptation \cite[LoRA]{hu2021lora}. 
LoRA efficiently captures changes in a weight matrix through a low-rank matrix, primarily to expedite the fine-tuning process. 
By encapsulating all significant training-induced changes within a low-rank matrix, we can quantify the similarity between two models through a direct comparison of the corresponding changes in their low-rank representation spaces.

We have validated the efficacy of RepMatch through a series of experiments across various NLP tasks, datasets, and models. The results demonstrate that LoRA matrices exhibit significant similarities among similar tasks, underscoring the reliability of RepMatch. Additionally, for each model, we identified a small yet representative subset within each dataset; models trained on these subsets consistently outperform those trained on comparably sized random subsets. 
In a definitive demonstration of cross-dataset utility, RepMatch successfully uncovers heuristics used in the automatic construction of the HANS challenge dataset.

\section{Related Work}

Quantifying the similarity between two distinct datasets is a well-researched topic. 
The theoretical concept of data similarity is linked to the traditional Kullback-Leibler divergence \citep{Kullback1951OnIA}, a non-symmetric measure that quantifies the dissimilarity between two probability distributions. 
Empirical measures like the Maximal Mean Discrepancy \cite[MMD]{Borgwardt2006IntegratingSB} have also been employed; MMD compares the means of samples drawn from two distributions in a high-dimensional feature space.

Building on the theoretical concepts, researchers have sought practical methods to estimate task difficulty and dataset similarity. 
\citet{tran2019transferability} utilized an information-theoretic approach to estimate task difficulty, demonstrating a strong correlation between their introduced hardness measure and empirical difficulty on transferability. 
A similar measure was proposed by \citet{alvarezmelis2020geometric} to quantify similarity between datasets.

Moving towards more empirical settings, \citet{hwang2020simex} presented a method to predict inter-dataset similarity using a set of pre-trained auto-encoders. 
Their approach involves inputting unknown data samples into these auto-encoders and evaluating the differences between the reconstructed outputs and the original inputs. 
While effective, this method requires additional computational resources and may be sensitive to the randomness inherent in the training environment.

Our method addresses these challenges by requiring no additional computation beyond regular fine-tuning with LoRA and remains robust to training randomness. 
Unlike previous approaches, RepMatch does not impose constraints on the size of the subsets it compares, which allows it to be categorized under data selection research. 
This flexibility contrasts with methods like that of \citet{swayamdipta-etal-2020-dataset}, who used training dynamics to divide a dataset into subsets of easy-to-learn, hard-to-learn, and ambiguous instances. 
Their method has limitations in analyzing individual instances or performing cross-dataset analysis.
The metric presented by \citet{ethayarajh2022understanding} can be used to quantify the complexity of individual instances relative to a specific distribution, which is useful for comparing datasets or subsets within a single dataset.  However, unlike RepMatch, this technique can not be used for comparisons of instances or segments across different datasets.

In the realm of data selection, a stream of prior research has aimed to find subsets of training examples that achieve performance close to training on the full dataset by using gradient information \citep{mirzasoleiman2020coresets, wang2021optimizing, yu2020gradient, killamsetty2021gradmatch}. Recently, \citet{xia2024less} proposed a method to estimate the influence function of a training data point to identify influential data in an instruction-tuning setting. While these methods focus on optimizing training efficiency, they may not directly address the comparison of dataset similarities.

\section{Methodology} \label{pm}

We introduce a method designed to assess the similarity between subsets of datasets, where subsets can be anything from individual instances to entire datasets. 
We define two subsets, \(\mathcal{S}_1 \) and \( \mathcal{S}_2 \), as similar if a model trained on \( \mathcal{S}_1 \) (denoted as \( \mathcal{M}_{\mathcal{S}_1} \)) exhibits a representation space akin to that of a model trained on \( \mathcal{S}_2 \) (\( \mathcal{M}_{\mathcal{S}_2} \)).
During standard fine-tuning, alterations to a specific weight matrix $\mathcal{W}_i$ (a specific weight matrix in layer $i$) are captured by $\Delta\mathcal{W}_i$, also known as the adaptation matrix. 
After fine-tuning, the updated model weights are then represented as $\hat{\mathcal{W}}_i = \mathcal{W}_i+ \Delta\mathcal{W}_i$.
These adaptation matrices are responsible for extracting task-specific features from the input and incorporating them into the pre-trained weight matrices. 
Since the pre-trained weights $\mathcal{W}_i$ remain constant across both models, comparing the representation spaces of \( \mathcal{M}_{\mathcal{S}_1} \) and \( \mathcal{M}_{\mathcal{S}_2} \) effectively boils down to analyzing the differences in \( \Delta \mathcal{W}_i\). 

The challenge in comparing models arises from the substantial size and high dimensionality of the weight matrices, particularly in modern language models. 
To manage this complexity, we propose using the LoRA method to encapsulate the \(\Delta \mathcal{W}_i\) matrices in a low-rank format.
In the following sections, we will provide a brief introduction to the LoRA method and explain how it facilitates our comparison of adaptation matrices between models.

\subsection{Background: LoRA}
\label{lora}

The goal of the LoRA method is to efficiently fine-tune a model $\mathcal{M}$ with a pre-trained weight matrix $\mathcal{W}_i$ on a specific dataset. LoRA achieves this by keeping the pre-trained weights ($\mathcal{W}_i$) frozen and allowing only the injected adaptation matrices ($\Delta \mathcal{W}_i$) to be updated during the fine-tuning process. 
To ensure parameter efficiency, LoRA restricts these $\Delta \mathcal{W}_i$ matrices to be low-rank. 
Specifically, if $\mathcal{W}_i$ is a ${d \times d}$ matrix, instead of updating this full-rank matrix directly, LoRA introduces two low-rank matrices  $(\mathcal{A}_i^r)_{d \times r}$ and $(\mathcal{B}_i^r)_{r \times d}$ for each layer $i$.
The product $\mathcal{A}_i^r\mathcal{B}_i^r$ then forms the adaptation matrix $\Delta \mathcal{W}_i^r$. 
While $\Delta \mathcal{W}_i^r$ retains the original $d \times d$ dimensions, its rank is limited to $r$, where $r<<d$, effectively reducing the number of parameters from $d^2$ to $2rd$. 

The authors of LoRA demonstrated that setting $r$ to be significantly smaller than $d$ does not generally result in substantial performance degradation across most NLP tasks (interestingly, they observed that in some cases, the performance of the model actually improves).

\subsection{Constraining Model Updates using LoRA}

While keeping the pre-trained weights frozen, we follow \citet{hu2021lora} and apply LoRA specifically to update attention matrices. Here, \( \Delta \mathcal{W}_i^r \) is formed by the product of $\mathcal{A}_i^r$ and $\mathcal{B}_i^r$--the LoRA matrices. Given the numerous possible combinations of  $\mathcal{A}_i^r$ and $\mathcal{B}_i^r$ that can result in the same \( \Delta \mathcal{W}_i^r\), our focus is solely on their resultant product, rather than on the individual matrices.

The low-rank nature of the $\Delta \mathcal{W}_i^r$ matrices in LoRA facilitates the efficient comparison of models. 
Since models trained on similar tasks are expected to extract analogous features, the LoRA matrices associated with a consistent pre-trained model should display similarities across comparable tasks and datasets. 
This insight drives our proposal to use these task-specific features, as identified by LoRA, to analyze both datasets and individual data instances.

Models \( \mathcal{M}_{\mathcal{S}_1} \) and \( \mathcal{M}_{\mathcal{S}_2}  \) are considered representationally similar if their corresponding LoRA matrices exhibit resemblance. 
Specifically, we compare the changes in the weight matrices, \(\Delta \mathcal{W}_i^r(\mathcal{M}_{\mathcal{S}_1} )\) and \(\Delta \mathcal{W}_i^r(\mathcal{M}_{\mathcal{S}_2} )\), across each layer $i$.
This method allows us to assess the similarity in their representation spaces by examining the modifications captured in these matrices.

\begin{figure*}[t!]
    \centering
    \subfloat[ \label{subfig1:a}]{
    \includegraphics[height=.35\textwidth,trim=10mm 0 10mm 0]{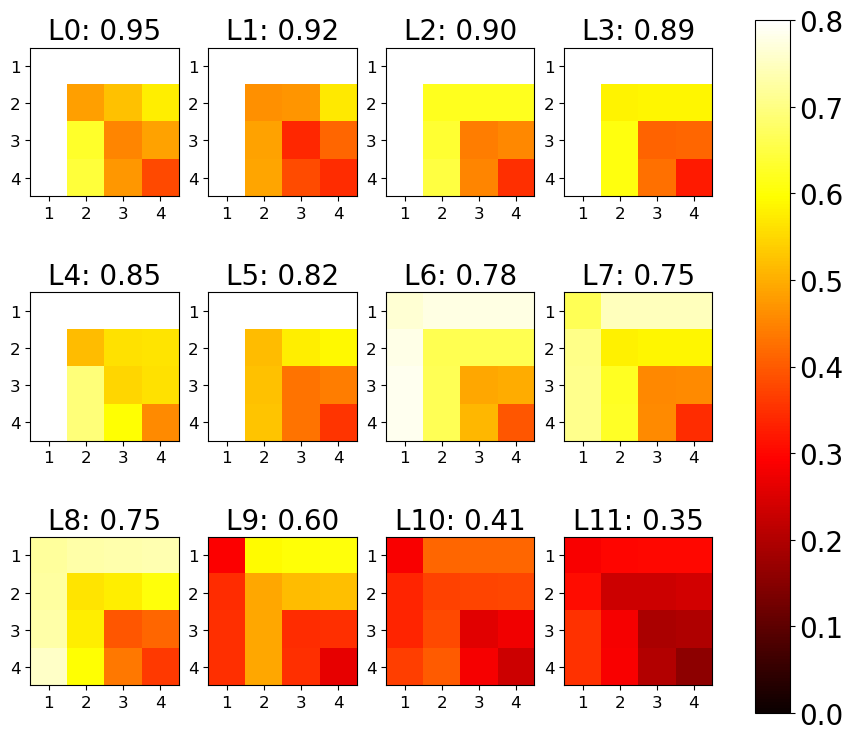}
    }\hspace{5mm}
    \subfloat[ \label{subfig1:b}]{
	\includegraphics[height=.35\textwidth,trim=10mm 0 10mm 0]{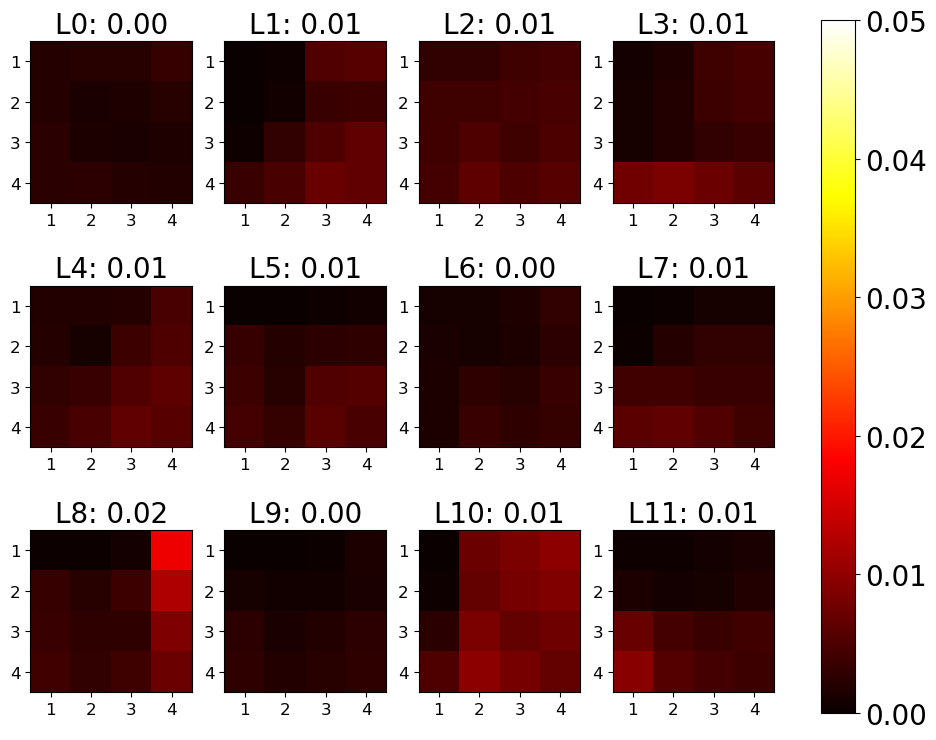}
	}
	\vspace{-3mm}
    \caption{Grassmann similarities of LoRA matrices (and the corresponding RepMatch scores on top) for
each of the 12 layers of two BERT\(_{base}\) models fine-tuned on SST-2 but (a) with different seeds, and (b) with random baseline (axes are $i$ and $j$ of the corresponding Grassmann similarity). Note the significantly different ranges of the two scales.}
    \label{fig:dataset-level}
    \vspace{-3mm}
\end{figure*}

\subsection{Grassmann Similarity}

To quantify the similarity of the subspaces formed by the two corresponding matrices from $\mathcal{M}_{\mathcal{S}_1} $ and $\mathcal{M}_{\mathcal{S}_2} $, we adopt the Grassmann similarity. 
The metric was used by \citet{hu2021lora} to discern subspace similarities across different ranks within the same dataset in order to verify the efficacy of low-rank matrices. 

Given two matrices, \(W^r\) and \(W^{r'}\), with ranks \(r\) and \(r'\) respectively, the Grassmann similarity measures the similarity between the subspaces they form as follows:
\begin{equation}
  \phi(\mathcal{W}^r, \mathcal{W}^{r^\prime}, i, j) = \frac{||U_{\mathcal{W}^r}^{i^\top} U_{\mathcal{W}^{r^\prime}}^j||_{F}^2}{\min(i, j)} \in [0,1]
  \label{eq:gras}
\end{equation}

\noindent where \(1 \leq i \leq r\) and \(1 \leq j \leq r'\), and both \(\mathcal{W}^r\) and \(\mathcal{W}^{r^\prime}\) are \(d \times d\) matrices. The matrix \(U\) is typically the right singular unitary matrix (obtained via SVD on \(\mathcal{W}\)), although the same can be achieved with left unitary matrix. 
The term \(U^j\) comprises the first \(j\) columns of \(U\), corresponding to the \(j\) largest singular values.

A high similarity implies that the subspace formed by the matrix of rank $r$ should predominantly reside within the subspace formed by $\mathcal{W}^{r^\prime}$. 
The matrices denoted by \(U\) can be interpreted as facilitating a change of basis. 
When these subspaces are in close proximity, the product of their corresponding \(U\) matrices tends toward unity, indicating a high degree of similarity between the subspaces. 
This proximity of subspaces is quantitatively expressed by the Grassmann similarity, which approaches zero as the alignment between the subspaces decrease.

\subsection{Computing RepMatch}

The RepMatch similarity score across the $i^{th}$ layers of two models $\mathcal{M}_{\mathcal{S}_1}$ and $\mathcal{M}_{\mathcal{S}_2}$ is computed as the Grassmann similarity between the corresponding adaptaion matrices
Note that, when comparing two matrices with rank \(r\), the Grassmann similarity produces an \(r \times r\) similarity matrix that reflects the similarity between all subspaces formed by the two matrices. 
The RepMatch score is computed as the highest value in this matrix. 
Also, the RepMatch score for the whole model is simply computed as the average of RepMatch scores across all layers (we leave other possible aggregation strategies to future work).  

In our initial experiments, we did not observe significant differences between various weight matrices in the attention block. Therefore, we chose to use value matrices for the remainder of our study.

Note that, when computing RepMatch between $\mathcal{M}_{\mathcal{S}_1}$ and $\mathcal{M}_{\mathcal{S}_2}$, the two models are actually the same pre-trained model, fine-tuned on two different subsets ${\mathcal{S}_1}$ and ${\mathcal{S}_2}$.
One can take the similarity between ${\mathcal{S}_1}$ and ${\mathcal{S}_2}$ as the similarity between the corresponding subsets ${\mathcal{S}_1}$ and ${\mathcal{S}_2}$ on which they are fine-tuned.
Therefore, in what follows, we will interchangeably use RepMatch as a measure of similarity between two models or two data subsets.

\section{Analysis Possibilities using RepMatch}
\label{sec:4}
The RepMatch similarity metric is unconstrained by the size or origin of the subsets, thus facilitating its application in a multitude of scenarios. 
For instance, it enables comparisons between individual instances and an entire dataset, or between subsets from distinct datasets.
In this section, we demonstrate the reliability of RepMatch for analysis in two of the possible scenarios: dataset-level and instance-level. 
To establish this, it is necessary to show that RepMatch is robust against the stochastic nature of the training environment. 
Here, the specific focus is on the training seed, verifying the extent to which RepMatch is affected by alterations in the initial point of the training.

In this section, we first demonstrate that RepMatch between two models trained on the same dataset with different random seeds is orders of magnitude higher than that of a model with partially random weights. 
Then, we show that the RepMatch between two models trained on a specific instance with different random seeds is higher than the average RepMatch between random instances of the same dataset.
For both these experiments, we used the {SST-2 sentiment analysis dataset \citep{socher-etal-2013-recursive}.}

\begin{figure*}[t!]
    \centering
    \subfloat[ \label{subfig2:a}]{
    \includegraphics[height=.35\textwidth,trim=10mm 0 10mm 0]{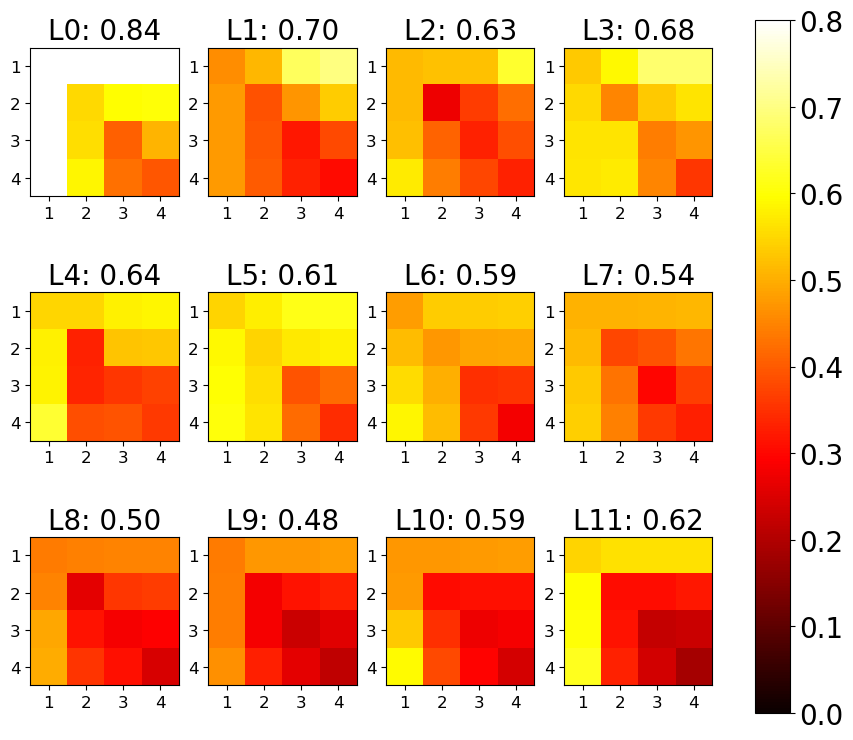}
    
    }\hspace{5mm}
    \subfloat[ \label{subfig2:b}]{
	\includegraphics[height=.35\textwidth,trim=10mm 0 10mm 0]{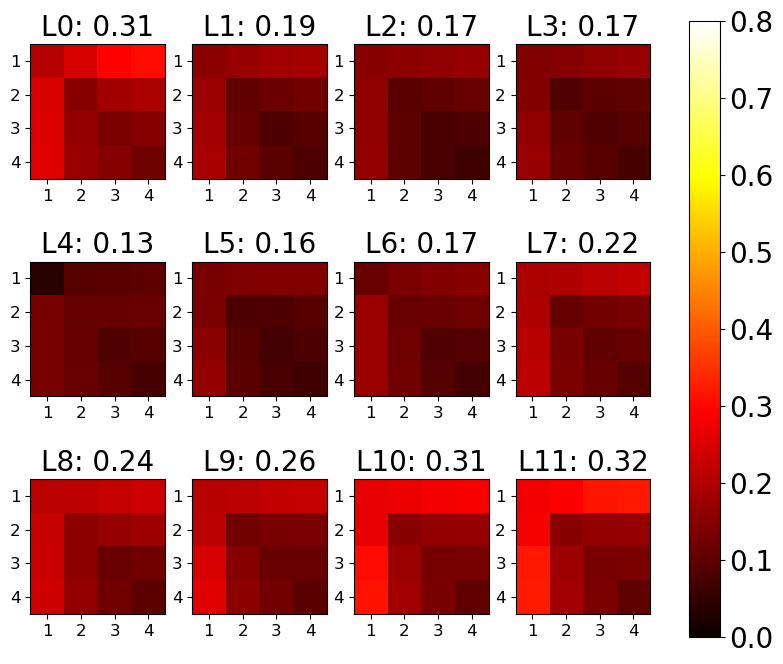}
	}
	\vspace{-3mm}
    \caption{Grassmann similarities of LoRA matrices (and the corresponding RepMatch scores on top) for each of the 12 layers of two BERT\(_{base}\) models: (a) fine-tuned on the same random instance from SST-2 but with different fine-tuning initial seeds, and (b) fine-tuned with the same seed but on two different instances from the dataset (axes are $i$ and $j$ of the corresponding Grassmann similarity). The results are the average of ten runs on different random seeds/instances.}
    \label{fig:instance-level}
    \vspace{-3mm}
\end{figure*}

\subsection{Dataset-level Analysis}

For dataset-level analysis, we consider the scenario where two identical models are fine-tuned on the same dataset under the same conditions, with the only difference being the random seed. 
We would expect these models to exhibit very similar characteristics.
Figure \ref{subfig1:a} shows the Grassmann similarities (and the corresponding RepMatch scores) 
for each layer $i$ of two BERT\(_{base}\) \citep{Devlin2019BERTPO} models fine-tuned on the dataset with different seeds. 
As can be seen from the heat maps, we used rank 4 for LoRA to obtain $\Delta \mathcal{W}_i$s).
Notably, there exists at least one vector in the corresponding matrix of each model that closely resembles its counterpart.

To demonstrate that the observed similarity is not due to chance, Figure \ref{subfig1:b} presents a random baseline for comparison. 
This figure compares the similarity of the fine-tuned model against itself, but with its entries shuffled.
This alteration creates a matrix that, while not drastically different, is distinct from one generated through a standard fine-tuning process. 
The average similarity score across different seeds exceeds 0.7, whereas for the baseline, the similarity score falls below 0.02, highlighting a significant difference.

Figure \ref{fig:dataset-level} also indicates that the largest similarity typically manifests in a single vector within each matrix. 
Consequently, setting the rank of LoRA matrices to one incurs minimal data loss. 
This is supported by the findings of \citet{hu2021lora}, which suggest that employing LoRA at a rank of one negligibly affects the model’s efficacy across many NLP tasks. 
For these reasons, and to efficiently compute the Grassmann similarity, we opted for rank of one in our experiments in Section \ref{5-3}

\subsection{Instance-level Analysis} \label{method:instance}

In order to validate if RepMatch scores are also robust with respect to random factors at the instance level, we carried out the following experiment.
Figure \ref{subfig2:a} shows the Grassmann similarities (and the corresponding RepMatch scores) between the two fine-tuned versions of the same BERT\(_{base}\) model on a randomly selected instance (the only varying factor being the different random seeds for fine-tuning). Numbers in the figure are averaged over 5 random instances.
Moreover, Figure \ref{subfig2:b} shows a similar experiment between two BERT\(_{base}\) models: one fine-tuned on the aforementioned instance and the other on a different random instance from the dataset but using the same training seed (selected 10 random instances from the dataset, Figure \ref{subfig2:b} highlights the average similarity).
It is evident that RepMatch between two models trained on the same instance but with different seeds is above 0.6, while for disparate instances, it hovers around 0.11.
These observations prove the robustness of RepMatch against random factors that can impact the model during fine-tuning.
Section \ref{exp} offers further empirical evidence supporting our proposed method.

\section{Experiments}
\label{exp}

In this section we show that datasets related to a specific task exhibit significantly higher RepMatch scores than datasets associated with unrelated tasks (Section \ref{sec:ds-sim}). 
We also carry out an experiment in which we isolate a small subset of \textit{representative} instances with the highest RepMatch scores (to the whole dataset). 
Notably, a model trained on this subset consistently outperforms one trained on a randomly selected subset of the same size (Section \ref{sec:inst-sim}). 
Finally, in Section \ref{sec:hans}, we show that RepMatch can be used across various datasets to identify heuristic patterns.

\subsection{Experimental Setup}
\label{sec:5.1}

\paragraph{Datasets.} We experimented with six dataset across three tasks: sentiment analysis (SST-2, SST-5, and IMDB \citep{maas-EtAl:2011:ACL-HLT2011}), textual entailment (MNLI \citep{N18-1101} and SNLI \citep{bowman-etal-2015-large}), and question answering (SQuAD v1 \citep{rajpurkar-etal-2016-squad}). 
 
\paragraph{Models and hyper-parameters.}
While the majority of experiments were conducted on BERT\(_{base}\), we also made additional trials on LLaMA2-7B \citep{touvron2023llama2openfoundation} and ELECTRA$_{base}$ \citep{clark2020electrapretrainingtextencoders} to verify that our findings are robust across different models.\footnote{Due to limited access to GPUs, we were constrained in our ability to test additional models, datasets, and configurations.}
All of these models were sourced from Hugging Face\footnote{\href{https://huggingface.co/models}{https://huggingface.co/models}}. 
Unless specified otherwise, our default fine-tuning setup involves integrating LoRA modules exclusively to the value matrices, while keeping all other model weights frozen. 
We employed a batch size of 40, conducting 10 epochs for sentiment analysis tasks and 5 epochs for other tasks. The rank of the LoRA matrices was set to one. For dataset-level analysis, we used a learning rate of $10^{-5}$, while instance-level experiments were conducted with a learning rate of $10^{-3}$ for speedup.
Due to limited resources, no hyper-parameter tuning was done for any of the settings.

\begin{figure}
    \centering
    \includegraphics[width= \columnwidth]{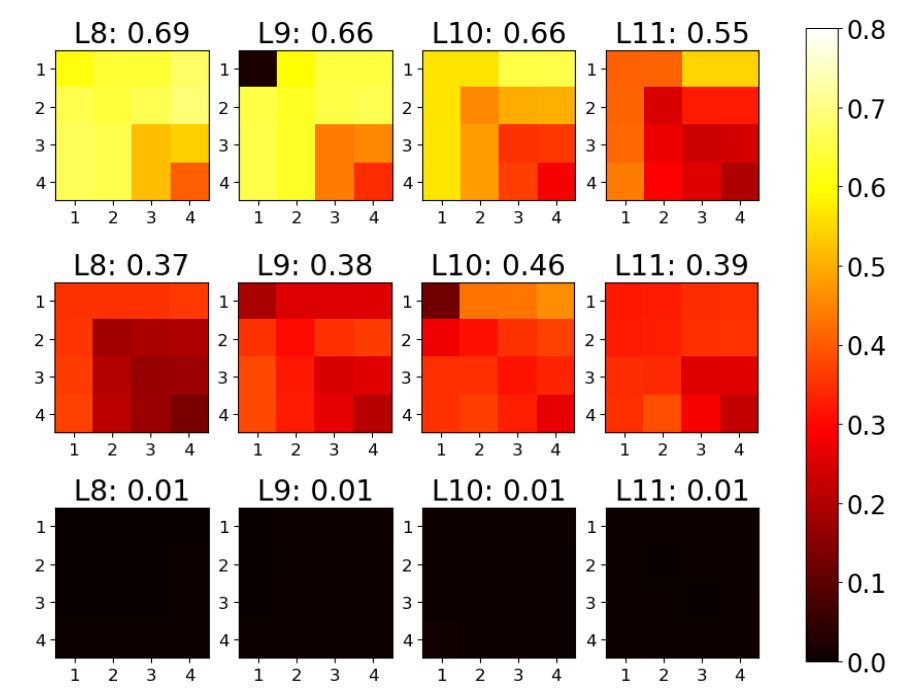}
    \caption{
    The Grassmann similarity (and the corresponding RepMatch scores on top) for three BERT\(_{base}\) models (axes are $i$ and $j$ in Grassmann similarity). The first row compares the last four layers of the a pre-trained BERT model fine-tuned on SST-2 and those of the same model fine-tuned on IMDB (see \ref{sec:appendix1} for all the layers).
    The other rows make similar comparisons across SST-2 and SST-5 (middle row) and SNLI (bottom row).
    As expected, the SST datasets (middle row) are the more similar. }
    \label{fig:imdb}
\end{figure}

\subsection{Dataset-level Similarity} 
\label{sec:ds-sim}

In Section \ref{sec:4}, we presented heat-maps to illustrate the similarities between subspaces created by the value matrix of a model trained on the same dataset but with differing training seeds. 
Also, we argued that tasks of a similar nature should exhibit comparable LoRA matrices (therefore high RepMatch scores).
To substantiate this claim, we conducted additional cross-dataset experiments.

Figure \ref{fig:imdb} shows the similarities between the SST-2 sentiment analysis dataset to three other datasets: IMDB and SST-5 for sentiment analysis (top and middle rows, respectively) and SNLI for entailment (bottom row).

Among these, SST-2 and SST-5 exhibit the highest similarity of 0.45, followed by the 0.26 score of SST-2 and IMDB, which are still associated to the same sentiment analysis task but originate from different sources.
In contrast, the RepMatch between SST-2 and MNLI is around 0.01, indicative of their distinct task natures.
Finally, the RepMatch between the SNLI and MNLI datasets was computed to be around 0.2, suggesting a closer relationship than with SST-2, yet highlighting considerable differences.
These results show that low-rank matrices encode valuable task-related features, which facilitate the comparison of subsets of instances.

\begin{table*}[t!]
\centering
\resizebox{\textwidth}{!}{%
\setlength{\tabcolsep}{9pt}
\begin{tabular}{lllllll} 
\toprule
                         \textbf{Dataset}    & \textbf{SST-2} & \textbf{SST-5} & \textbf{IMDB} & \textbf{MNLI} & \textbf{SNLI} & \textbf{SQuAD} 
\\ \midrule
Random     & 69.53 $\pm${\small 2.31}  & 35.25 $\pm${\small 0.24}  & 70.29 $\pm${\small1.73}  & 58.09 $\pm${\small 1.50} & 38.34 $\pm${\small 0.38} &    23.44 $\pm${\small 3.39}            \\
RepMatch   & 81.03 $\pm${\small 1.78}   & 39.59 $\pm${\small 1.25}  & 78.70 $\pm${\small 1.89} & 61.79 $\pm ${\small 0.10} & 43.29 $\pm ${\small 0.19} &  25.12 $\pm${\small 1.87}         
 \\ \bottomrule
\end{tabular}%
}
\caption{\label{table:bert-100}
The accuracy performance of BERT when fine-tuned on two subsets of 100 instances, selected randomly (first row) or based on their representativeness according to RepMatch. The results shown are average $\pm${\small std} over 3 runs (for SQuAD, F1 scores are reported).
}
\end{table*}

\paragraph{Representation-based baseline.}

We were curious to see if the representational similarity between instances could serve as a proxy for their similarity from the model's perspective, reflecting the knowledge encoded within the model.
To compare two datasets, $\mathcal{D}_1$ with $n$ instances and $\mathcal{D}_2$ with $m$ instances, based on their representations, we constructed a similarity matrix $\mathcal{C}$. 
Each element $\mathcal{C}_{i,j}$ in this matrix is the cosine similarity between the \texttt{[CLS]} representations of the $i^{th}$ instance of $\mathcal{D}_1$ and the $j^{th}$ instance of $\mathcal{D}_2$, derived from a pre-trained BERT$_{base}$ model.\footnote{Given the matrix's symmetry, only entries for $i \geq j$ or $j \geq i$ needed to be calculated.}
The overall similarity was quantified by calculating the percentage of cosine similarities that exceeded a threshold (for this experiment, we opted for the high similarity of 0.9). 
For larger datasets like SNLI and MNLI, the similarity matrix becomes impractically large. To address this, random subsets of 10,000 instances were chosen from each dataset for comparison. For datasets with fewer than 10,000 instances, the entire dataset was used.

Table \ref{table:baseline} compares the similarity of SNLI to four other datasets, as calculated using the proposed baseline and RepMatch.
In addition to the datasets introduced in Section \ref{sec:5.1}, we experimented with the semantic textual similarity task, specifically the STS-B dataset\footnote{\href{https://huggingface.co/datasets/mteb/stsbenchmark-sts}{https://huggingface.co/datasets/mteb/stsbenchmark-sts}}.
This dataset has a similar two-sentence format as SNLI and MNLI entailment datasets (but focuses on the semantic similarity of the pair of sentences rather than their inference relationship).
Although SNLI and STS-B are associated with distinct tasks, the \texttt{[CLS]} representations perceive them as being highly similar. 
In contrast, RepMatch identifies them as two entirely different datasets. 
This suggests that RepMatch can better capture the instances according to the knowledge they carry, rather than structural or topical features (which seem to have been captured by the \texttt{[CLS]} representations of the pre-trained model).

\begin{table}[!]
\centering
\resizebox{\columnwidth}{!}{%
\setlength{\tabcolsep}{6pt}
\begin{tabular}{lrrrrr} 
\toprule
                         \textbf{Dataset}   & \bf SNLI & \bf MNLI & \bf HANS & \bf SST-2 & \bf STS-B 
\\ \midrule
Baseline     & 91.4~  & 91.1~~ & 89.7~~ & 4.8~~ & 92.4~~         \\
RepMatch    & 69.7~ & 19.9~~ & 5.6~~ &  1.0~~ & 0.8~~  
 \\ \bottomrule
\end{tabular}%
}
\caption{{\label{table:baseline}
The similarity between SNLI and four other datasets (and SNLI itself) according to the representation-based baseline and RepMatch. The entries represent the percentage similarity calculated using each method. A random subset of 10,000 instances was used for comparison in all datasets except STS-B, which has only 8,628 instances.}
}
\end{table}

\subsection{Instance-level Similarity} \label{5-3}
\label{sec:inst-sim}

Thanks to the flexibility of RepMatch, one can use the metric for identifying the more \textit{representative} instances of a dataset. 
We consider an instance $x \in \mathcal{X}$ to be representative if the RepMatch between $x$ and $\mathcal{X}$ (the entire dataset) is high.
Accordingly, the most representative instances in the dataset $\mathcal{X}$ are those with the highest RepMatch scores to $\mathcal{X}$.

The process of calculating individual RepMatch scores involves running a pre-trained model with a batch size of one to update the LoRA matrices. The updated model is then compared to a model previously fine-tuned on the entire dataset. To ensure a fair comparison, the model is reset to its original pre-trained state before processing each subsequent instance.

Table \ref{table:bert-100} compares, across different datasets, the BERT model fine-tuned on the 100 most representative instances against that fine-tuned on a randomly selected subset of the same size. 
The results clearly indicate a consistent performance improvement for the RepMatch-based instances.
According to Table \ref{table:bert-100}, SST-2 and IMDB exhibit the most significant gap. 
We attribute this to the limited matrix rank, which might be less restricting for simpler tasks. Increasing the matrix rank could potentially enhance this disparity across other datasets, albeit possibly hitting a performance ceiling.\footnote{The results reported in the table are based on full fine-tuning. However, the experiment was also repeated with LoRA, resulting in a performance decline of 3 to 5 percent for both Random and RepMatch groups, yet the gap between them largely remained the same.}

Additionally, we experimented with two other models, LLaMA2-7B and ELECTRA$_{base}$, to verify the effectiveness of RepMatch across different models. 
The experiments on LLaMa were particularly time-intensive due to the model's size.
Hence, we opted for SST-5 only (fine-tuned using LoRA), reporting 4\% improvement in the performance when using the RepMatch group (0.30 vs 0.34) compared to the random one.

As for ELECTRA$_{base}$, we observed around 5\% improvement on IMDB (90.44 vs. 85.5) and around 8\% on SST-2 (88.12 vs. 80.7).

\begin{figure}
    \centering
    \includegraphics[width= \columnwidth]{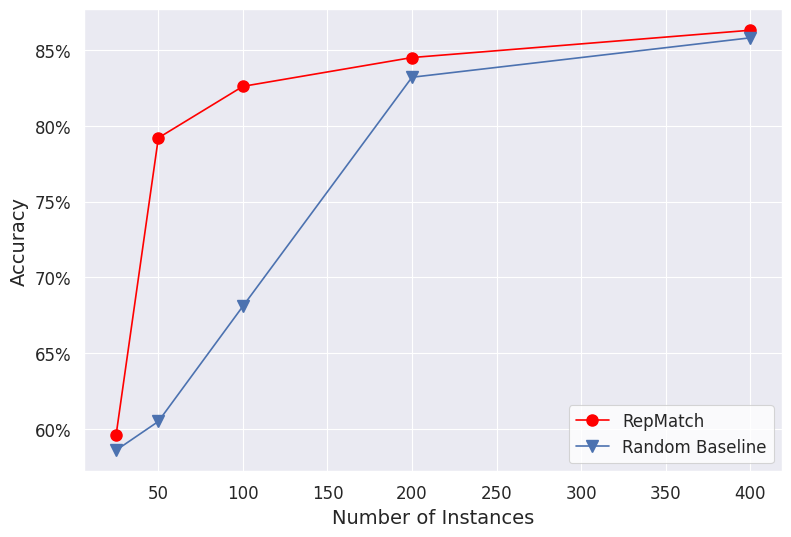}
    \caption{
    Performance variation of a BERT\(_{base}\) model, fine-tuned on different subset sizes of SST-2. The top line is for the subset of most representative instances, selected using RepMatch, whereas the other line is for the randomly chosen subset.
    }
    \label{fig:learning-curve}
\end{figure}

\paragraph{The impact of subset size.}
To verify the impact of subset size, we carried out experiments with varying subset sizes using the SST-2 and BERT\(_{base}\) model. 
As depicted in Figure \ref{fig:learning-curve}, any subset smaller than 400 selected using RepMatch can consistently outperform a randomly selected subset of the same size, although the performance gap decreases.

\subsubsection{Detecting out-of-distribution instances}
\label{sec:hans}

The RepMatch method has no limitations on the size or domain of the considered set, thus making it applicable in various analytical contexts. 
To demonstrate this, an experiment was designed to showcase the cross-dataset capabilities of the method.
Specifically, we opted for detecting out-of-distribution instances.

Previous studies have identified certain superficial artifacts in widely used textual entailment datasets, such as MNLI and SNLI \cite{gururangan-etal-2018-annotation,rajaee-etal-2022-looking}.
It has been argued that models tend to leverage these artifacts (which usually arise as a result of decisions made during dataset construction) to achieve high performance without necessarily learning the task. 
One such artifact in textual entailment datasets is that high overlap between the premise and the hypothesis is likely indicative of the entailment label.

\begin{table}
\centering
\resizebox{\columnwidth}{!}{%
\setlength{\tabcolsep}{16pt}
\begin{tabular}{lrrr} 
\toprule
                         \textbf{Dataset}    & \textbf{Full} & \textbf{Mid} & \textbf{No}  
\\ \midrule
MNLI     & 1,016  & 46,569   & 8,629           \\
SNLI   & 949  & 53,873 & 21,396        
 \\ \bottomrule
\end{tabular}%
}
\caption{\label{table:NLI-sets}
The distribution of instances across different subsets in the two NLI datasets.
The sets are extracted based on the degree of overlap between the premise and the hypothesis. The ``Full'' set encompasses instances with full overlap, the ``Mid'' set contains instances where the overlap between the premise and the hypothesis ranges from 60\% to 80\%, and the ``No'' set, as the name suggests, includes instances where there is no overlap. All sets have non-entailment label.}

\end{table}

\begin{table*}[!]
\centering
\resizebox{\textwidth}{!}{%
\setlength{\tabcolsep}{8pt}
\begin{tabular}{lllllll} 
\toprule
                         \textbf{Dataset}    & \textbf{SST-2} & \textbf{SST-5} & \textbf{IMDB} & \textbf{MNLI} & \textbf{SNLI} & \textbf{SQuAD}
\\ \midrule
RepMatch$_{r=1}$   & 80.44 $\pm${\small 1.73}   & 39.59 $\pm${\small 1.25}  & 77.85 $\pm${\small 2.10} & 50.83 $\pm ${\small 1.44} & 38.19 $\pm ${\small 0.53}   & 21.89 $\pm ${\small 3.83}               \\
RepMatch$_{r=4}$  & 82.46 $\pm${\small 1.20}   & 35.65 $\pm${\small 1.12}  & 85.66 $\pm${\small 3.68} & 62.50 $\pm ${\small 0.25} & 38.70 $\pm ${\small 1.10} & 24.36 $\pm ${\small 2.57}      
 \\ \bottomrule
\end{tabular}%
}
\caption{\label{table:rank4}
Accuracy performance of BERT fine-tuned on two subsets of 100 instances, selected based on their representativeness according to RepMatch with a rank of 1 (first row) or 4. The results shown are average $\pm${\small std} over 3 runs (for SQuAD, F1 scores are reported). A random subset of 25,000 instances was used for each dataset, except for SST-5, which has only 8544 instances.
}
\end{table*}

\begin{table}
\centering
\resizebox{\columnwidth}{!}{%
\setlength{\tabcolsep}{7pt}
\begin{tabular}{lccc} 
\toprule
                         \textbf{Dataset}    & \textbf{Full} & \textbf{Mid} & \textbf{No}  
\\ \midrule
MNLI     & {76.18$\pm${\small4.30}}  & 72.25$\pm${\small 1.11} & 70.24$\pm${\small 1.64}           \\
SNLI   & {78.24$\pm${\small3.78}}  & 69.37$\pm${\small 1.97} & 73.62$\pm${\small 2.73}        
 \\ \bottomrule
\end{tabular}
}
\caption{\label{table:heuristics}
The RepMatch scores between each subset of the two NLI datasets and the HANS dataset. 
The values are the average over three random subsets, computed using BERT\(_{base}\).}
\end{table}

To highlight this, challenge sets like HANS \cite{mccoy2019right} were created to test the models’ genuine understanding of the task. This dataset includes examples that counter the heuristics in the NLI datasets. For instance, in the case of overlap bias, a high overlap between the premise and the hypothesis results in a non-entailment label in HANS, contrasting with MNLI and SNLI. Henceforth, we will refer to these two datasets as NLI datasets.

We hypothesize that non-entailment instances in the training set of NLI datasets with high overlaps will be more similar to the HANS dataset than other instances.
To validate our hypothesis, we leveraged the dataset-level analysis setting. 
The difference is that here we measure the similarity across datasets, i.e., between a subset of the NLI datasets and the entire set of HANS instances.
To achieve this, we extracted three sets from each NLI dataset--all with non-entailment labels but different degrees of overlap between premise and hypothesis: full overlap, 60\% to 80\% overlap, and no overlap. 
Table \ref{table:NLI-sets} shows the number of instances in each set. 
Given the varied set sizes, 300 samples were randomly selected from each overlap subset (without replacement) for a fair comparison. 

The RepMatch for each set was computed using BERT\(_{base}\) with respect to a model trained on HANS. 
The experiment was repeated three times on different subsets, the average similarity of which is reported in Table \ref{table:heuristics}.\footnote{For brevity, all scores were multiplied by \(\frac{1}{\text{learning rate}}\), which does not affect the comparison.}
As expected, the set containing full overlap instances with non-entailment labels showed the highest average similarity to the HANS dataset.
This demonstrates that the RepMatch method can be used to find or analyze bias or heuristics with respect to another dataset, which could be useful for out-of-distribution generalization purposes.

\section{The Effect of Rank}

As explained in Section \ref{lora}, the LoRA method constrains the weight update matrices, \(\Delta \mathcal{W}_i^r\)s, to have a low rank of \(r\). In Section \ref{5-3}, the rank was set to 1. Here, the instance-level experiments were repeated with a rank of 4 instead. Since computing RepMatch with higher ranks requires significantly more computational time, a random subset of 25,000 instances was used for each dataset. Table \ref{table:rank4} shows the results of fine-tuning a pre-trained model on the 100 instances with the highest RepMatch scores.
We observe that increasing the rank generally improves results across datasets, with the exception of SST-5. 
This may be attributed to the imbalance in labels among the top-100 instances with the highest RepMatch scores.

In the case of SST-5, this is unsurprising, as the dataset itself is inherently imbalanced. 
When selecting the 100 most representative samples with rank 4, the majority belonged to the dominant label class, in contrast to sampling with rank 1 which was more balanced.
These results require further scrutiny since the experiments were conducted on a small portion of the entire dataset. We leave further analysis to future work.

\section{Conclusion}

In this study, we approached the problem of dataset analysis from a unique perspective. We proposed a method to identify similarities between subsets of training instances by examining the similarities within the representation space of models trained on them. We overcame the challenges of complexity and heavy parameters of language models by utilizing the LoRA method to constrain changes in the representation space.
Although we employed LoRA, alternative parameter-efficient fine-tuning methods \cite{liu2024doraweightdecomposedlowrankadaptation, akbartajari-etal-2022-empirical} that limit weight updates might be beneficial in this setting.

Our findings suggest that RepMatch can be employed to compare tasks and datasets, conduct instance-level analysis to discover heuristics in a dataset, and perform subset analysis to identify a smaller subset that achieves reasonable performance and outperforms a randomly selected subset of the same size.
The experiments demonstrated that the proposed method can be utilized in a variety of situations and is not limited by the size of the subset or its domain.
RepMatch proves useful for comparing tasks and datasets, conducting instance-level analysis to uncover dataset heuristics, and identifying high-performing subsets. 
We hope the technique opens new avenues for analysis of datasets and models, from the new viewpoint of knowledge captured by a model from a training instance.

\section*{Limitations}
In the instance-level setting, the relationship between instances within a training batch is not taken into account. There exists a possibility that a model might exhibit better performance when trained with two less representative instances in a batch, rather than two highly similar ones. This presents a potential avenue for enhancing the experimental setup. 

Furthermore, while we demonstrated that the entire dataset and individual instances are robust to the random seed of the training environment, the randomness of training and instances in a batch can have a non-negligible effect.

The majority of our experiments were conducted on BERT\(_{base}\), with a few experiment on LLaMA2 and ELECTRA$_{base}$. Due to GPU limitations, further experiments were not viable. Although our focus was on Transformer models with a textual modality and our evaluations were based on three different classification tasks, we believe this method is applicable to other modalities and settings.

\bibliography{citations}
\bibliographystyle{acl_natbib}

\appendix
\section{Appendix}
\label{sec:appendix}
\subsection{Grassmann Similarities and Corresponding RepMatch}
\label{sec:appendix1}
\begin{figure*}[!]
    \centering
    \subfloat[ \label{sst2_sst5}]{
    \includegraphics[height=.35\textwidth,trim=10mm 0 10mm 0]{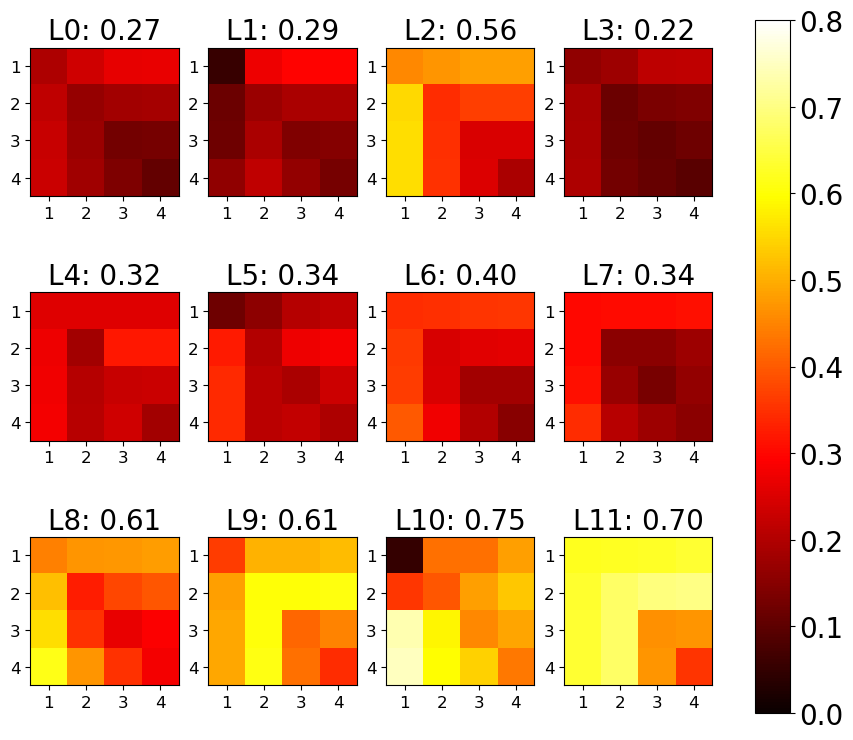}
    }\hspace{5mm}
    \subfloat[ \label{sst2_imdb}]{
	\includegraphics[height=.35\textwidth,trim=10mm 0 10mm 0]{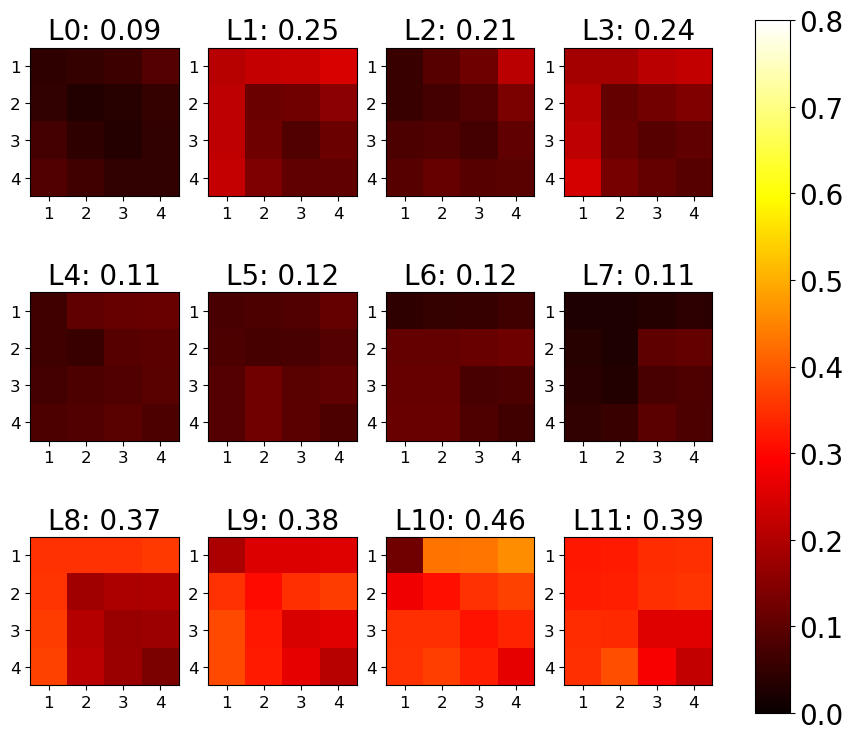}
	}
	\vspace{-3mm}
    \caption{Grassmann similarities of LoRA matrices (and the corresponding RepMatch scores on top) for
each of the 12 layers of two BERT\(_{base}\) models fine-tuned on SST-2 and (a) SST-5 / (b) IMDB (axes are $i$ and $j$ of the corresponding Grassmann similarity).}
    \label{fig:5}
    \vspace{-3mm}
\end{figure*}

\begin{figure*}[!]
    \centering
    \subfloat[ \label{sst2_snli}]{
    \includegraphics[height=.35\textwidth,trim=10mm 0 10mm 0]{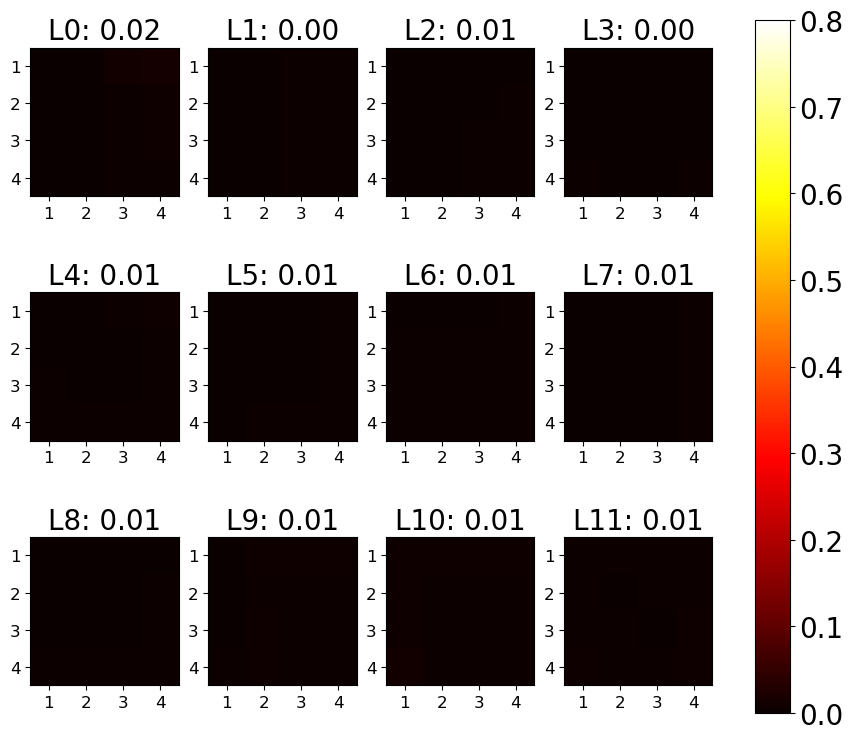}
    }\hspace{5mm}
    \subfloat[ \label{mnli-snli}]{
	\includegraphics[height=.35\textwidth,trim=10mm 0 10mm 0]{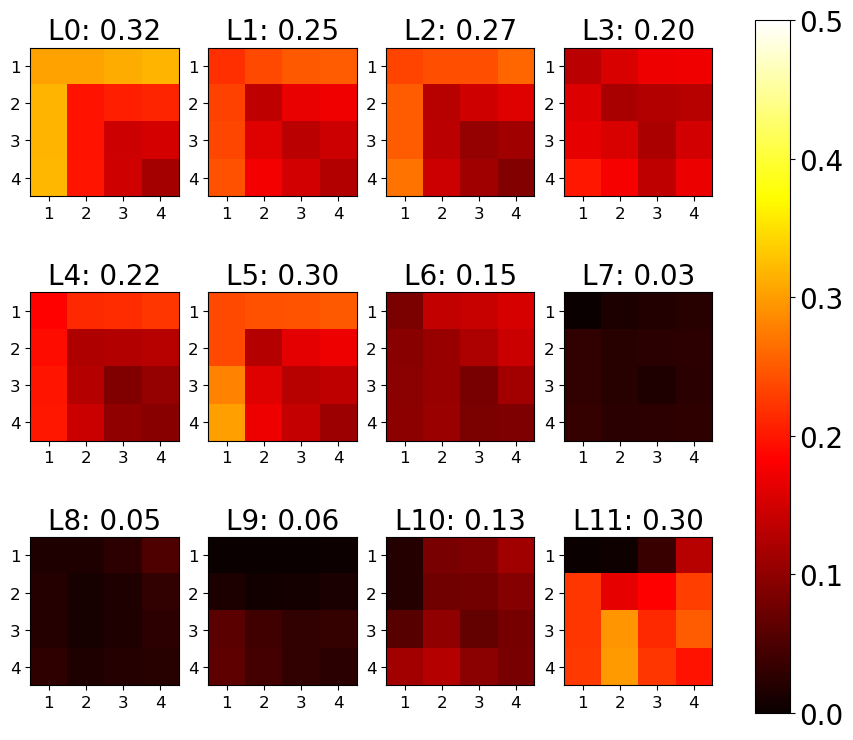}
	}
	\vspace{-3mm}
    \caption{Grassmann similarities of LoRA matrices (and the corresponding RepMatch scores on top) for
each of the 12 layers of two BERT\(_{base}\) models fine-tuned on SNLI and (a) SST-2 / (b) MNLI (axes are $i$ and $j$ of the corresponding Grassmann similarity).}
    \label{fig:6}
    \vspace{-3mm}
\end{figure*}

 Here we provide full heat-maps representing Grassmann similarity  and RepMatch between SST2 and SST5/ IMDB/ SNLI (Figures \ref{fig:5} and \ref{sst2_snli}). The same for comparing SNLI and MNLI is also provided in Figure \ref{mnli-snli}.

\end{document}